# FOREGROUND SEGMENTATION BASED ON MULTI-RESOLUTION AND MATTING

*Xintong Yu[1,2], Xiaohan Liu[1,2], Yisong Chen[1]*

[1] Graphics Laboratory, EECS Department, Peking University
[2] Beijing University of Posts and Telecommunications

**ABSTRACT**

We propose a foreground segmentation algorithm that does foreground extraction under different scales and refines the result by matting. First, the input image is filtered and resampled to 5 different resolutions. Then each of them is segmented by adaptive figure-ground classification and the best segmentation is automatically selected by an evaluation score that maximizes the difference between foreground and background. This segmentation is upsampled to the original size, and a corresponding trimap is built. Closed-form matting is employed to label the boundary region, and the result is refined by a final figure-ground classification. Experiments show the success of our method in treating challenging images with cluttered background and adapting to loose initial bounding-box.

## 1. INTRODUCTION

Foreground segmentation plays an important role in image analysis and computer vision. Popular approaches are mainly based on graph, probability, information, or variational theories [1, 2, 3]. Super-pixel generation often acts as an effective pre-processing for foreground extraction due to its advantages in information transfer and computational efficiency [4, 5]. In particular, patches generated by the mean-shift algorithm [4] are better described statistically compared to other super-pixel generators [6]. Various matting techniques and systems have been proposed to extract high quality mattes from both still images and video sequences [7]. Matting is also a commonly used post-processing method for foreground segmentation to remove boundary artifacts [8]. Multi-hypothesis based segmentation draws increasing attention recently due to its capacity of obtaining better results by robustly fusing different candidates [9, 10].

By joint use of adaptive mean-shift and multiple hypotheses fusion, the recently proposed adaptive figure-ground classification method [11] (abbreviated as f-g classification) achieves great success in foreground extraction from natural images, particularly for foregrounds

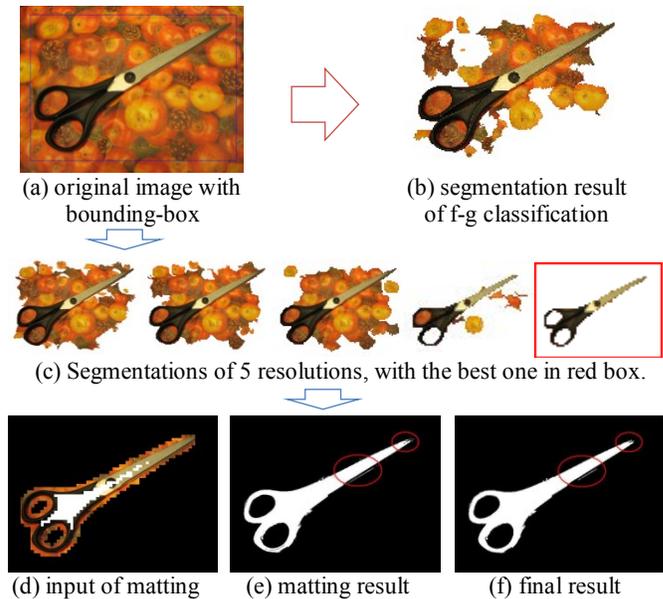

(a) original image with bounding-box  (b) segmentation result of f-g classification

(c) Segmentations of 5 resolutions, with the best one in red box.

(d) input of matting  (e) matting result  (f) final result

**Figure.1.** Blue arrows show the pipeline of our algorithm. As the comparison, (b) shows the result of f-g classification. Images of (c) are adjusted to the same size regardless of their resolution. Note the slight differences between (e) and (f) within the red ellipses (zoom in for details).

with irregular contours. However, this method suffers from loose bounding-box or cluttered background due to the limitation of the mean-shift over-segmentation. It may produce too many patches within the region of interest, which causes poor background prior estimate and thus degrades the performance. The authors of [11] suggested employing bigger bandwidths to circumvent these cases. However, there lacks theoretically sound guide about how to set the parameters and only empirical values are attempted. This drawback may cause failures when treating difficult scenes with cluttered backgrounds, as shown in Fig.1 (b).

To solve this problem, we propose a segmentation algorithm based on multi-resolution and closed-form matting. First, we filter and resample the input image to 5 different resolutions. Then we use f-g classification to generate a segmentation candidate for each resolution (see Fig.1 (c)) and choose the best one according to the maxmin-cut score. After the selected segmentation is upsampled to its original size, the boundary of its foreground and background becomes an area (see Fig.1 (d)). We use closed-form matting to classify the pixels of this area into the foreground or background category (see Fig.1 (e)). In the end, we do f-g classification again on the matting result to filter out tiny fragments and get final segmentation (see Fig.1 (f)).

The advantage of our approach is twofold. First, it can effectively treat cluttered images. Second, it is more robust to loose initial bounding-box.

## 2. FOREGROUND SEGMENTATION BASED ON MULTI-RESOLUTION AND CLOSED-FORM MATTING

In the proposed foreground extraction method, two main processes are consecutively involved: 1) the multi-resolution segmentation, 2) the closed-form matting post-processing.

### 2.1. The multi-resolution segmentation

Images often have different levels of details at different scales [12]. A series of studies have been carried out on using multi-resolution methods to help segmentation or contour detection [13]. The work of [12] shows that, when an image is reduced, the cluttered texture of its background will be correspondingly smoothed. That is, the edges of the cluttered patterns become unobservable while the edges of the main objects are still salient.

The f-g classification method fails to treat cluttered images mainly because the super-pixel generation step may produce too many patches. We treat this difficulty by considering the segmentation on a proper smaller scale. To find the proper scale, we use the multi-resolution method to generate multiple candidates for selecting.

In particular, we downsample the image to 1/2, 1/4, 1/6, 1/8 and 1/10 of its original size, and use f-g classification to obtain a segmentation result for each resolution. When a cluttered image is reduced, the number of patches will decrease, and the patches may be classified more correctly to foreground or background category. In general, segmentation results of the 5 resolutions are largely different, and there often exists some good results within them (the 5$^{th}$ resolution for Fig.1). Fig.2 and Tab.1 show more examples of multi-resolution segmentation and the corresponding number of patches. Images (b) and (d) get satisfactory results in the 3$^{rd}$ resolution, whereas image (a) and (c) fail to reach a good result until the 5$^{th}$ resolution.

We do not use pyramid style [14] that only takes 2^n resolutions here, since the sizes smaller than 1/10 usually produce too few patches for a normal classification routine and those larger than 1/2 are mostly too large for smoothing details. In practical, we find that the 5 sizes in Table 1 are sufficiently good for most cases. Sometimes only the first few resolutions are used (see Fig.2 (b)), depending on whether the patches generated under the given resolution is enough to do f-g classification.

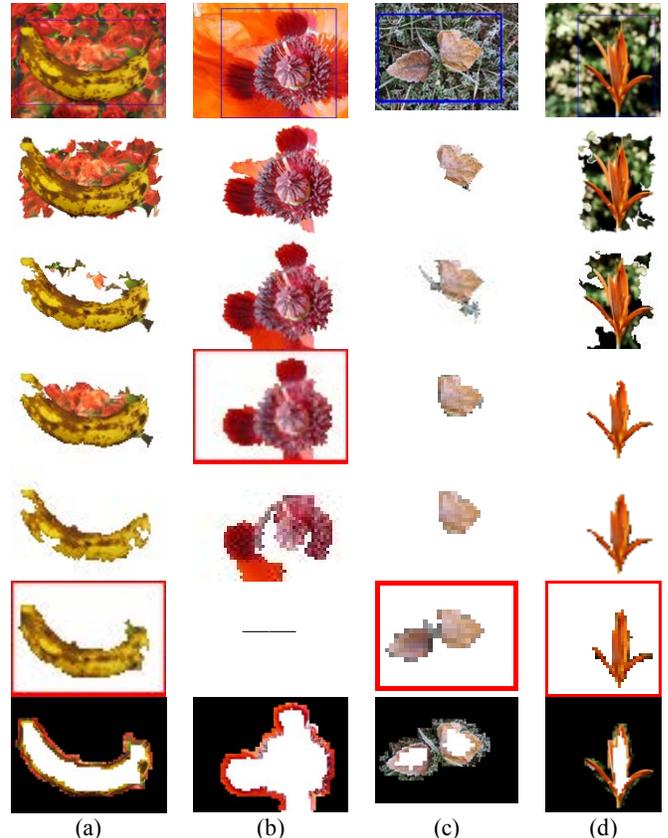

(a)  (b)  (c)  (d)

**Figure.2.** From top to bottom: the original image; the segmentation results of 5 resolutions with the automatically selected one in the red box; the trimap of matting. (All images are adjusted to the same size regardless of their resolution.)

| resolutions | 1/2 | 1/4 | 1/6 | 1/8 | 1/10 |
|---|---|---|---|---|---|
| Figure 1 | 266 | 92 | 45 | 22 | 14 |
| Figure 2 (a) | 239 | 74 | 39 | 23 | 14 |
| Figure 2 (b) | 114 | 40 | 18 | 10 | — |
| Figure 2 (c) | 86 | 23 | 11 | 5 | 3 |
| Figure 2 (d) | 157 | 44 | 19 | 12 | 10 |

**Table.1.** Number of patches of the 5 resolutions of Fig.1 and 2.

Given the 5 segmentation candidates for an input image, we need a score function to automatically choose the best. For cluttered scenes, the segmentation under the original size is often too conservative and has bigger background regions. So we adopt a more aggressive score, the maxmin-cut (abbreviated as m-cut), to counteract such tendency. M-cut tends to select smaller foreground region in comparison to average-cut and sum-cut [11]. With the m-cut score, we have a better chance of correcting the bias of the segmentation result of the original resolution.

In particular, with 5 candidate segmentations {S1, S2, S3, S4, S5} at 5 different scales, the selected one is given by

$$S_{m-\text{cut}} = \arg\max_{i=1..5} D(F(S_i), B(S_i)), \quad (1)$$

where $F(S_i)$ and $B(S_i)$ are respectively the foreground and background groups in the final segmentation map of the i-th candidate, and $D(F, B)$ is the minimum distance between foreground and background regions under a given segmenta-

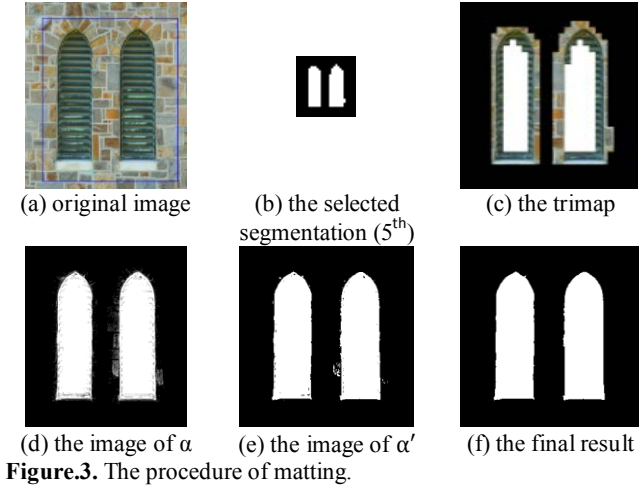

(a) original image  (b) the selected segmentation (5th)  (c) the trimap

(d) the image of α  (e) the image of α′  (f) the final result

**Figure.3.** The procedure of matting.

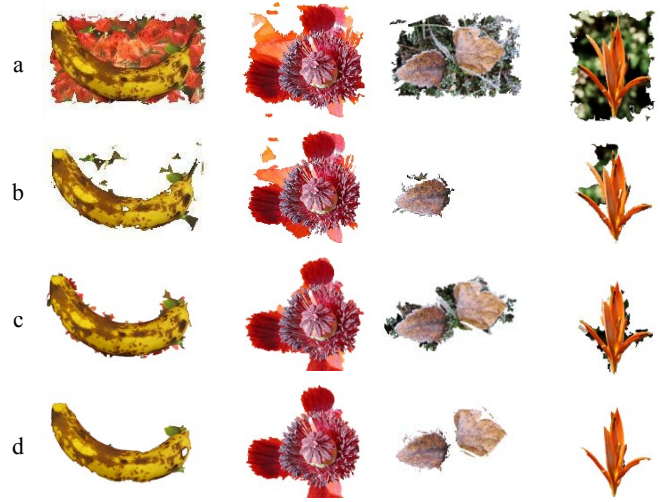

**Figure.4**. Comparison of results of different algorithms (Original images in Figure 2).
a: Results of f-g classification
b: Results of refined f-g classification
c: Results of multi-resolution + f-g
d: Results of multi-resolution + matting

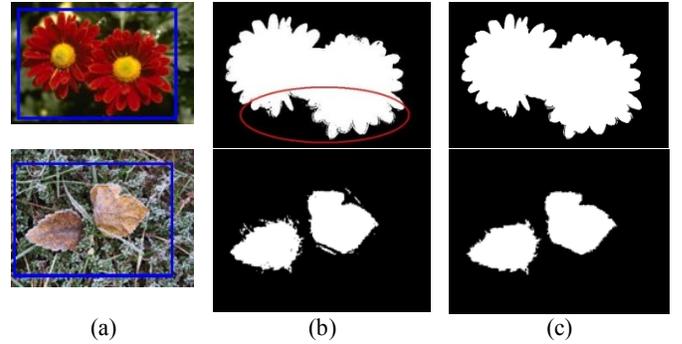

(a)  (b)  (c)

**Figure.5.** Comparison of 2 post-processing methods. (a) input images, (b) results of multi-resolution + matting, (c) results of running f-g after (b). Red circle shows the differences.

tion [11].

The m-cut score works well in a broadly applicable environment. In rare cases where the user is not satisfied with the selection of m-cut, results of other resolutions can be manually selected to do the matting process followed.

### 2.2 The closed-form matting post-processing

In the last step, we get the optimal segmentation under a certain resolution (see Fig.3 (b)). After upsampled to the original size, its boundary becomes an area, which is defined as the "unknown region". This makes a trimap that can be well treated by the technique of matting (see Fig.1 (d), the 7th row of Fig.2, and Fig.3 (c)). We take closed-form matting as our post-processing method due to its flexibility in modeling and computation [15]. The algorithm is briefly described as below.

In image matting methods, the color of the i-th pixel is assumed to be a linear combination of the corresponding foreground and background colors F and B:

$$I_i = \alpha_i F_i + (1 - \alpha_i) B_i, \quad (2)$$

where $\alpha_i$ is the pixel's foreground opacity. Under the local smoothness assumption that each F or B is a linear mixture of two colors over a small window (3×3 or 5×5), the $\alpha$ values in a small window $w$ can be expressed as:

$$\alpha_i \approx \sum_c a^c I_i^c + b, \ \forall i \in w, \ \left(a^c = \frac{1}{F-B}, b = -\frac{B}{F-B}\right). \quad (3)$$

The closed-form matting derives a cost function from local smoothness assumptions:

$$J(\alpha, a, b) = \sum_{j \in I} \left( \sum_{i \in w_j} \left(\alpha_i - \sum_c a_j^c I_i^c - b_j\right)^2 + \varepsilon \sum_c a_j^{c^2} \right). \quad (4)$$

From (4), a quadratic cost function in α is obtained, and the optimal α of each pixel can be solved in closed form via solving a sparse linear system (see Fig. 3 (d)). Then we classify the i-th pixel ($p_i$) of the unknown region as:

$$\alpha_i \begin{cases} \geq 0.5 \to \alpha'_i = 1, \ p_i \in F \\ < 0.5 \to \alpha'_i = 0, \ p_i \in B \end{cases} \quad (5)$$

This gives all the pixels a binary label of 1 or 0, corresponding to foreground or background (see Fig. 3 (e)). After all foreground pixels are extracted from the image, we run f-g classification again to filter out the tiny fragments generated by matting and get the final result (see Fig. 3 (f)).

Aside from the approach above, we also tried two other methods to refine the segmentation obtained from multi-resolution. The first is to run f-g classification algorithm on the original image, using the outer boundary line of the trimap as the bounding-curve instead of the user-specified rectangle. It turns out that, although the bounding-curve is tighter than the bounding-box, the image of the original size may still be too cluttered for f-g classification to deal with. The extracted objects are often accompanied with small background fragments, as shown in Fig. 4 (c).

The second is to run matting on the trimap without further refinement. Generally, the results are satisfactory (see Fig.1 (e) and Fig.4 (d)). However, sometimes there still exist tiny misclassified fragments or undesired ghosting around extracted objects (see Fig. 3 (e), Fig. 5 (b)).

Our proposed strategy can effectively handle the above drawbacks by joint use of matting and f-g classification. That is, we run f-g classification after matting, using the matting result to define the background mask. This time most fragments are classified correctly and most boundary ghosting are removed (see Fig. 3 (f) and Fig. 5 (c)). The differences between Fig.3 (e) and (f), and Fig.5 (b) and (c) show the effect of the last step of f-g classification. Experiments show that this post-processing strategy of cascading matting and f-g classification obtains the most accurate segmentation (see section 3).

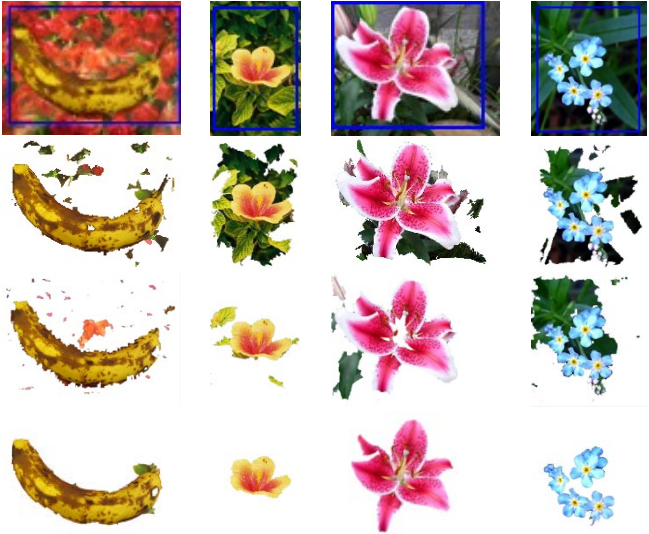

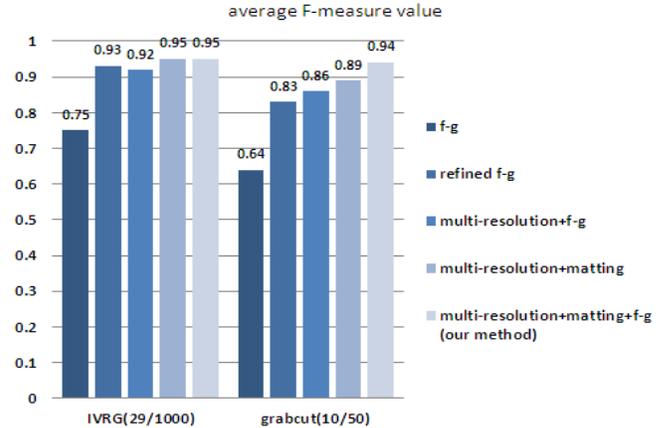

**Figure.6.** Segmentation results from loose bounding-boxes. From top to bottom are: original images, results from refined f-g classification, results from grabcut, and results from our method.

## 3. EXPERIMENTS

We take cluttered images from various sources to evaluate the performance of the proposed method. In Fig.4, we compare some example segmentations of our method and f-g classification. The f-g classification algorithm fails to provide accurate segmentation for cluttered images (see Fig.4 (a)). The authors of [11] suggested employing bigger bandwidths to treat these cases. We call it "the refined f-g classification". It can improve the result to some extent, but may also cause bad result if the bandwidths are not appropriate (see Fig. 4 (b)). Our method outperforms both the original and the refined f-g classification (see Fig. 4 (d)).

The robustness to loose initial bounding-boxes is another advantage of employing multi-resolution. Fig. 6 shows that our method performs better under loose bounding-box in comparison to the refined f-g classification and grabcut, which are more sensitive to the position and tightness of the bounding-box.

Finally, we choose some cluttered images from the grabcut dataset [16] and the IVRG dataset [17] for F-measure evaluation [11]. In particular, we first run the mean-shift algorithm on all images, and select those with more than 300 patches within the region of interest as cluttered ones. We respectively find 10 and 29 out of 50 and 1000 images from the 2 datasets, which makes our test sets. Fig.7 compares the average F-measure values of 5 segmentation strategies. We can see that the strategy of combining multi-resolution, matting and f-g classification achieves the highest F-measure value of over 0.94 in both datasets.

**Figure. 7.** Performance comparison on 2 datasets of 5 strategies.

## 4. CONCLUSION

We propose a foreground segmentation algorithm based on multi-resolution and matting. The image is first filtered and resampled to 5 resolutions. Then f-g classification is used to treat each resolution and generate segmentation candidates. The best one is automatically selected by m-cut. Next, closed-form matting is employed to label the boundary region and the result is refined by a final f-g classification. Our method achieves success in segmenting cluttered images and adapting to loose bounding-boxes.

## 5. RELATION TO PRIOR WORK IN THE FIELD

Our work has focused on segmenting challenging images with cluttered background, using the strategy of combining multi-resolution, matting and f-g classification. The f-g classification and closed-form matting algorithms are respectively cited from [11] and [15]. The method of multi-resolution is related to recent studies on the scale of edges [12, 13].